\documentclass[11pt,twocolumn]{article}

\usepackage[dvips]{graphicx}

\begin{document}
\date{}

\title{\Large\bf Learning to automatically detect features for mobile
  robots using second-order Hidden Markov Models \footnote{also in \cite{aycard04a}}}

\renewcommand{\thefootnote}{$\fnsymbol{footnote}$}

\author{\small \begin{tabular}{ccc} 
Olivier Aycard & Jean-Fran{\c c}ois Mari & Richard Washington\footnotemark[1]\\ 
GRAVIR - IMAG & LORIA & Autonomy and Robotics Area \\ 
Joseph Fourier University & Nancy2 University & NASA Ames Research Center\\ 
38 000 Grenoble, France & 54 506 Vand{\oe}uvre Cedex, France & Moffett Field, CA 94035, USA\\ 
Olivier.Aycard@imag.fr & jfmari@loria.fr & richw@email.arc.nasa.gov\\ \end{tabular}}

\maketitle


\bibliographystyle{plain}

\footnotetext[1]{NASA contractor with RIACS.}

\renewcommand{\thefootnote}{\arabic{footnote}}

\vspace{2 mm} 
\noindent
{\em 
{\bf Abstract:} In this paper, we propose a new method based on Hidden Markov
Models to interpret temporal sequences of sensor data from mobile
robots to automatically detect features.
Hidden Markov Models have been used for a long time in 
pattern recognition, especially in speech recognition. Their main
advantages over other methods (such as neural networks) are their
ability to model noisy temporal signals of variable length. 
We show in this paper that this approach is well suited for
interpretation of temporal sequences of mobile-robot sensor data. We
present two distinct experiments and results: the first one in an
indoor environment where a mobile robot learns to detect features like
open doors or T-intersections, the second one in an outdoor
environment where a different mobile robot has to identify 
situations like climbing a hill or crossing a rock.

}
\vspace{1 mm}

\section{Introduction}

A mobile robot operating in a dynamic environment is provided with
sensors (infrared sensors, ultrasonic sensors, tactile sensors,
cameras\ldots) in order to perceive its environment. Unfortunately,
the numeric, noisy data furnished by these sensors are not directly
useful; they must first be interpreted to provide accurate and usable
information about the environment. This interpretation plays a crucial
role, since it makes it possible for the robot to detect pertinent
features in its environment and to use them for various tasks.

For instance, for a mobile robot, the automatic recognition of features is an important issue for the following reasons:
\begin{enumerate}
\item For successful navigation in large-scale environments, mobile
robots must have the capability to localize themselves in their
environment. Almost all existing localization
approaches \cite{Borenstein96} extract a small set of features. During
navigation, mobile robots detect features and match them with known
features of the environment in order to compute their position;
\item Feature recognition is the first step in the automatic
construction of maps. For instance,
at the topological level of his
``spatial semantic hierarchy'' system, Kuipers \cite{Kuipers00} incrementally
builds a topological map by first detecting pertinent features while
the robot moves in the environment and then determining the link
between a new detected feature and features contained in the current
map;
\item Features can be used by a mobile robot as subgoals for
a navigation plan~\cite{Lazanas95}.
\end{enumerate}

In semi-autonomous or remote, teleoperated robotics, automatic
detection of features is a necessary ability.  In the case of limited
and delayed communication, such as for planetary rovers, human
interaction is restricted, so feature detection can only be
practically performed through on-board interpretation of the sensor
information.  Moreover, feature detection from raw sensor data,
especially when based on a combination of sensors,
is a complex task that generally cannot be done
in real time by humans, which would be necessary even if teleoperation
were possible given the communication constraints.
For all these reasons, feature detection has received considerable
attention over the past few years. This problem can be classified with
the following criteria:
\paragraph{Natural/artificial} The first criterion is the nature of the feature. The features
can be artificial, that is, added to the existing environment. Becker et al~\cite{Becker95} define a set of artificial
features\footnotemark[2] located on the ceiling and use a camera to
detect them.\footnotetext[2]{The features are patterns composed of 3x3
squares, and each square is colored in black or white} Other techniques
use natural features, that is, features already existing in the environment. For instance, Kortenkamp, Baker, and Weymouth~\cite{Kortenkamp92} use
ultrasonic sensors to detect natural features like open doors and
T-intersections.

Using artificial features makes the process of detection and
distinction of features easier, because the features are designed to
be simple to detect.  But this approach can be time-consuming,
because the features have to be designed and to be positioned in the
environment. Moreover, using artificial features is impossible in
unknown or remote environments.

\paragraph{Analytical/statistical methods}  Feature detection has been addressed by different approaches such as
analytical methods or pattern classification methods. In the analytical
approach, the problem is studied as a reasoning process.
A knowledge based system uses rules to build a
representation of features. For instance, Kortenkamp, Baker, and Weymouth~\cite{Kortenkamp92} 
use rules about the variation of the sonar sensors to learn
different types of features and adds visual information to distinguish
two features of the same type. 
In contrast, a statistical pattern-classification system attempts to
describe the observations coming from the sensors as a random process.
The recognition process consists of the association of the signal
acquired from sensors with a model of the feature to identify. For
instance, Yamauchi~\cite{Yamauchi95} uses ultrasonic sensors to build evidence
grids~\cite{Elfes89}. An evidence grid is a grid corresponding to a
discretization of the local environment of the mobile robot. In this
grid, Yamauchi's method updates the probability of occupancy of each
grid tile with several sensor data. To perform the detection, he defines an
algorithm to match two evidence grids.

These two approaches are complementary. In the analytical approach, we aim
to understand the sensor data and build a representation of
these data. But as the sensor data may be noisy, so their interpretation
may not be straightforward; moreover, overly simple descriptions of
the sensor data (e.g., ``current rising, steady, then falling'') may
not directly correspond to the actual data.

In the second approach, we build models that represent the statistical
properties of the data. This approach naturally takes into account the
noisy data, but it is generally difficult to understand the correspondence
between detected features and the sensor data.

A solution that combines the two approachs could build
models corresponding to human's understanding of the sensor data, and
adjust the model parameters according to the statistical properties of
the data.


\paragraph{Automatic/manual feature definition.} The set of features to detect could be given manually or
discovered automatically~\cite{Thrun98}. 
In the manual approach, the set is defined
by humans using the perception they have of the environment. Since
high level robotic system are generally based loosely on human
perception, the integration of feature detection in such a system is
easier than for automatically-discovered features. Moreover, in
teleoperated robotics, where humans interact with the robot, 
the features must correspond to the high level perception of
the operator to be useful. These are the main reasons the set is almost always
defined by humans.  However, properly defining the features so that they
can be recognized robustly by a robot remains a difficult problem; this
paper proposes a method for this problem.
In contrast, when features are discovered automatically, humans must find
the correspondence between features perceived by the robot and features they
perceive. The difficulty now rests on the shoulders of the humans.


\paragraph{Temporally extended/instantaneous features.}
Some features can only be identified by considering a temporal
sequence of sensor information, not simply a snapshot, especially with
telemetric sensors. Consider for example the detection of a
feature in~\cite{Kortenkamp92} or the construction of an evidence grid
in~\cite{Yamauchi95}: these two operations use a temporal
sequence of sensor information. In general, instantaneous (i.e.,
based over a simple snapshot) detection is less robust than
temporal detection.


This paper describes an approach that combines an analytical approach
for the high-level topology of the environment with a statistical
approach to feature detection.  
The approach is designed to detect natural, temporally extended features
that have been manually defined.
The feature detection uses Hidden
Markov Models (HMMs). HMMs are a particular type of probabilistic
automata. The topology of these automata corresponds to a human's
understanding of sequences of sensor data characterizing a particular
feature in the robot's environment.  We use HMMs for pattern
recognition. From a set of training data produced by its sensors and
collected at a feature that it has to identify --- a door, a rock,
\ldots --- the robot adjusts the parameters of the corresponding model
to take into account the statistical properties of the sequences of
sensor data.  At recognition time, the robot chooses the model whose
probability given the sensor data --- the {\it a posteriori}
probability --- is maximized.  We combine analytical methods to define
the topology of the automata with statistical pattern-classification
methods to adjust the parameters of the model.

The HMM approach is a flexible method for handling the large
variability of complex temporal signals; for example, it is a standard
method for speech recognition~\cite{Rabiner89}.  In contrast to
dynamic time warping, where heuristic training methods for estimating
templates are used, stochastic modeling allows probabilistic and
automatic training for estimating models.
The particular approach we use is the second-order HMM (HMM2), which
have been used in speech recognition \cite{Mari97}, often out-performing
first-order HMMs.

This paper is organized as follow. We first define the HMM2 and
describe the algorithms used for training and
recognition. Section~\ref{sec:application-robotics} is the description
of our method for feature detection combining HMM2s with a grammar-based
analytical method describing the environment. In
section~\ref{sec:first-experiment}, we present an experiment of
our method to detect natural features like open doors or
T-intersections in an indoor structured environment for an autonomous
mobile robot. A second experiment on a semi-autonomous mobile robot in
an outdoor environment is described in
section~\ref{sec:second-experiment}. Then we report related work in
section~\ref{sec:related-work}. We give some conclusions and
perspectives in section~\ref{sec:conclusion}.

\section{Second-order Hidden Markov Models}
\label{sec:hmm2}
In this section, we only present second-order Hidden Markov Models
in the special case of multi dimensional continuous observations
(representing the data of several sensors). We also detail the
second-order extension of the learning algorithm (Viterbi
algorithm) and the recognition algorithm (Baum-Welch algorithm). A
very complete tutorial on first order Hidden Markov Models can be
found in Rabiner~\cite{Rabiner89}.

\subsection{Definition}
In an HMM2, the underlying state sequence is a second-order Markov chain. 
Therefore, the probability of a transition between two states at time $t$
depends on the states in which the process was at time $t-1$ and $t-2$. 

A second order Hidden Markov Model $\lambda$ is specified by:

\begin{itemize}
\item a set of $N$ states called S containing at least one final state;
\item a 3 dimensional matrix {$a_{ijk}$} over S x S x S
\begin{eqnarray}
a_{ijk} = Prob(q_t = s_k / q_{t-1} = s_j, q_{t-2} = s_i) \\ \nonumber
= Prob(q_t = s_k / q_{t-1} = s_j,q_{t-2} = s_i, \\
q_{t-3} = ...  ) \nonumber
\end{eqnarray}
with the constraints 
\begin{displaymath}
 \sum_{k=1}^N a_{ijk} = 1 ~~~ with   ~ 1 \leq i \leq N ~,~ 1 \leq j \leq N
\end{displaymath}
where  $q_t$ is the actual 
state at time $t$ ;

\item each state $s_i$ is associated with a mixture of Gaussian distributions :
\begin{equation}
b_i(O_t) = \sum_{m=1}^M c_{im}  {\cal N}(O_t; {\mu}_{im}, { \Sigma}_{im}),
\end{equation}
\begin{displaymath}
  ~~~ with  ~~~ \sum_{m=1}^M c_{im} = 1
\end{displaymath}
where $O_t$ is the input vector (the frame) at time t. The mixture of
Gaussian distributions is one of the most powerful probability
distribution to represent complex and multi-dimensional probability
space. 
\end{itemize}
The probability of the state sequence 
\[
Q = q_1,q_2,...,q_T
\]
is defined as 
\begin{equation}
Prob(Q) = 
        \pi_{q_1}  a_{q_1q_2} \prod_{t=3}^{T} a_{q_{t-2}q_{t-1}q_t}
\end{equation}
where $\Pi_{i}$ is the probability of state $s_i$ at time $t = 1$ and 
$a_{ij}$ is the probability of the transition $s_i \to s_j$ at
time $t = 2$.

Given a sequence of observed vectors 
$O~=~o_1,o_2,...,o_T$, 
the joint state-output
probability $Prob(Q,O/\lambda)$,  is defined as :
\begin{eqnarray}
Prob(Q,O/\lambda) &= \Pi_{q_1} b_{q_1}(O_1) a_{q_1q_2}b_{q_2}(O_2) \times
\\ \nonumber
& 
  \prod_{t=3}^{T} a_{q_{t-2}q_{t-1}q_t} b_{q_t}(O_t). \label{eq:alig}
\end{eqnarray}

\subsection{The Viterbi algorithm}
The recognition is carried out by the Viterbi algorithm 
\cite{Forney73} which determines the most likely state sequence given a sequence of observations.

In Hidden Markov Models, many state sequences may generate the 
same
observed sequence $O~=~o_1,...,o_T$. Given one such output sequence, we are 
interested in determining the most likely
state sequence $Q~=~q_1, ...,q_T$  that could have generated
the observed sequence.  

The extension of the Viterbi algorithm to HMM2 is straightforward. 
We simply replace the reference to a state in the state space {\bf S}
 by a reference to an element of the 2-fold product space  {\bf S~x~S}.
The most likely state sequence is found by using the probability
of the partial alignment ending at transition $(s_j,s_k)$ at times $(t-1,t)$. 
\begin{eqnarray}
\delta_t(j,k) &=&  Prob(q_1,...q_{t-2},  \\ \nonumber
 &~& q_{t-1}=s_j,q_{t}=s_k,\\ \nonumber
&~& o_1,...,o_t/\lambda) \\ \nonumber
 &~&2 \leq t \leq T ,~~ 1 \leq j, k \leq N. 
\end{eqnarray}
Recursive computation is given by equation
\begin{eqnarray}
\delta_t(j,k) = max_{1 \leq i \leq N }[\delta_{t-1}(i,j) \cdot a_{ijk}]
\cdot b_k(O_t) \\ \nonumber
3 \leq t \leq T, ~~1 \leq j, k \leq N.
\end{eqnarray}

The Viterbi algorithm is a dynamic programming search that computes the best 
partial state sequence up to time $t$ for all states. 
The most likely state sequence $q_1,...,q_T$ is obtained by keeping
track of back pointers for each computation of which previous
transition leads to the maximal partial path probability. By tracing
back from the final state, we get the most likely state sequence.

\subsection{The Baum-Welch algorithm}
The learning of the models is performed by the Baum-Welch algorithm using the 
maximum likelihood estimation criteria that determines the best model's 
parameters according to the corpus of items. Intuitively, this algorithm counts
 the number of occurrences of each transition 
between the states and the number of occurrences of each observation in a given state in the training corpus. Each count is weighted by
the probability of the alignment (state, observation). It must be noted that 
this criteria does not try to 
separate models like a neural network does, but only tries to increase the 
probability that a model generates its corpus independently of what the other 
models can do.

Since many state sequences may generate a given output sequence, the 
probability that a model $\lambda$ generates a sequence $o_1$,...,$o_T$ is
given by the sum of the joint probabilities (given in
equation~\ref{eq:alig}) over
all state sequences (i.e, the marginal density of output sequences).
To avoid combinatorial explosion, a recursive computation similar
to the Viterbi algorithm can be used to evaluate the above sum. The
forward probability $\alpha_t(j,k)$ is :
\begin{eqnarray}
\alpha_{t+1}(j,k) = prob( & O_1,..., O_t = o_1,...,o_t, \\ \nonumber
& q_{t-1} = s_j, q_t = s_k / \lambda ).
\end{eqnarray}
This probability represents the probability of starting from state 0
and ending with the transition ($s_j$, $s_k$) at time t and generating
output $o_1$,...,$o_t$ using all possible state sequences in between. The
Markov assumption allows the recursive computation of the forward
probability as :
\begin{eqnarray}
{\alpha_{t+1}(j,k) =  \sum_{i=1}^{N}{\alpha}_{t}(i,j).a_{ijk}.b_k(O_{t+1}),~~}
\\ \nonumber
{2 \leq t \leq T-1,~~1 \leq j, k \leq N}
\end{eqnarray}
This computation is similar to Viterbi decoding except that summation
is used instead of max. The value $\alpha_T(j,k)$ where $s_k$ = N is the
probability that the model $\lambda$ generates the sequence $o_1,...,o_t$.
Another useful quantity is the backward function $\beta_t(i,j)$, 
defined as the probability of the partial observation sequence from $t+1$
to $T$, given the model $\lambda$ and the transition $(s_i,s_j)$ 
between times $t-1$ and $t$, can be expressed as :
\begin{eqnarray}
\beta_t(i,j) &=& Prob(O_{t+1},...O_T/
\\ \nonumber 
&~& q_{t-1}=s_i,q_t=s_j,\lambda), \\ \nonumber
&~& 2 \leq t \leq T-1, ~~ 1 \leq i,j \leq N.
\end{eqnarray}
The
Markov assumption allows also the recursive computation of the backward
probability as :
\begin{enumerate}
\item Initialization
\[
\beta_T(i,j) ~=~ 1 ~~\textrm {if}~~ 1 \leq i,j \leq N
\]
\item Recursion for $ 2 \leq t \leq T-1 $ 
\begin{eqnarray}
{\beta_t(i,j) ~=~  \sum_{i=1}^{N}{\beta}_{t+1}(j,k).a_{ijk}.b_k(O_{t+1})}
\\ \nonumber 
{1 \leq i,j \leq N}
\end{eqnarray}
\end{enumerate}

Given a model $\lambda$ and an observation sequence $O$,
we define  ${\eta}_t(i,j,k)$ as the probability of the  transition 
$s_i \longrightarrow s_j \longrightarrow s_k$ 
between  $t-1$ and $t+1$ during the emission of the observation
sequence. 
\begin{displaymath}
{\eta}_t(i,j,k) = \mathrm{P}(q_{t-1} = s_i, q_{t} =
s_j,q_{t+1} = s_k /O,\lambda),
\end{displaymath} 
\begin{displaymath}
 ~~ 2 \leq t \leq T-1. 
\end{displaymath} 
We deduce:
\begin{equation} \label{eq:ETA}
{\eta}_t(i,j,k)= 
\frac{{\alpha}_t(i,j) a_{ijk}  b_k(O_{t+1}){\beta}_{t+1}(j,k)}{P(O|{\lambda})},
\end{equation}
\begin{displaymath}
~~ 2 \leq t \leq T-1.
\end{displaymath}

As in the first order, we define 
${\xi}_t(i,j)$ and $\gamma_t(i)$:
\begin{equation}
{\xi}_t(i,j) = \sum_{k=1}^{N}{\eta}_{t} (i,j,k),
\end{equation}
\begin{equation}
{\gamma}_t(i) =  \sum_{j=1}^{N}{\xi}_t (i,j).
\end{equation}

${\xi}_t(i,j)$ represents the {\em aposteriori} probability that the
stochastic process accomplishes the transition $s_i \to s_j$
between $t-1$ and  $t$ assuming the whole utterance.

${\gamma}_t(i)$  represents the  {\em aposteriori} probability that the 
process  is in the state  $i$ at time $t$ assuming the whole utterance.

At this point, to get the new maximum likelihood estimation (ML) of the
$HMM_2$, we can choose two ways of normalizing: one way gives an
$HMM_1$, the other an $HMM_2$.

The  transformation in $HMM_1$ is done by averaging the counts   
${\eta}_{t}(i,j,k)$ over all the states $i$ that have been visited
at time  $t-1$. 

\begin{equation} \label{normalisation_HMM1}
{\eta}_{t}^1(j,k) = \sum_{i=1}^N {\eta}_{t}(i,j,k)
\end{equation}
is the classical first order count of  transitions between  2 $HMM_1$ states
between  $t$ and $t+1$. 

Finally, the first-order maximum likelihood (ML) estimate of
$\overline{a_{ijk}}$  is:
\begin{equation} 
{\overline{a_{ijk}}} 
=
\frac{
{\sum_t{ 
{\eta}_{t}^1(j,k)}}
}
{\sum_{k,t}{{\eta}_{t}^1(j,k)}}
= 
{\frac{\sum_{i,t} {\eta}_{t}(i,j,k)}
{\sum_{i,k,t} {\eta}_{t}(i,j,k)}}.
\label{eq_norm1}
\end{equation}
This value is independent of  $i$ and can be written as
$\overline{a_{jk}}$.

The second-order ML estimate of   $\overline{a_{ijk}}$ 
is given by the equation:
\begin{eqnarray}
\overline{a_{ijk}} &=& 
{\frac{\sum_{t} {\eta}_{t}(i,j,k)} 
{\sum_{k,t} {\eta}_{t}(i,j,k)}} \nonumber \\
&=&
\frac{ \sum_{t=1}^{T-2} {\eta}_{t+1}(i,j,k)}
{\sum_{t=1}^{T-2}{\xi}_t(i,j)}. 
\end{eqnarray}

The ML estimates of the mean and covariance are given by the formulas:
\begin{eqnarray}
\overline{{\mu}_i} &=& \frac{ \sum_t {\gamma}_t(i) O_t} 
{\sum_t{\gamma}_t(i)}, \label{eq:mu} \\
\overline{{\Sigma}_i} &=&  \frac{ \sum_t {\gamma}_t(i) 
 (O_t -{\mu}_i) (O_t -{\mu}_i)^t}
{\sum_t{\gamma}_t(i)}. \label{eq:sigma}
\end{eqnarray}

\section{Application to mobile robotics}
\label{sec:application-robotics}
The method presented in this paper performs feature detection by
combining HMM2s with a grammar-based description of the environment.
To apply second order Hidden Markov Models to automatically detect
features, we must accomplish a number of steps.  In this section we
review these steps and our approach for treating the issues arising
in each of them.  In the following sections we expand further on
the specifics for each experiment.

The steps necessary to apply HMM2s to detect features are the following:
\begin{enumerate}
\item Defining the number of distinct features to identify and their 
characterization.

As Hidden Markov Models have the ability to model signals whose
properties change with time, we choose a set of sensors (as the
observations) that have noticeable variations when the mobile robot is
observing a particular feature. The features are chosen for the fact
that they are repeatable and human-observable (for the purposes of
labeling and validation). So, we define coarse rules to identify
each feature, based on the
variation of the sensors constituting the observation to identify
each feature.  These rules are for human use, for segmentation and labeling
of the data stream of the training corpus. The set of
chosen features is a complete description of what the mobile robot can
see during its run. All other unforeseen features are treated as
noise.

    
\item Finding the most appropriate model to represent a specific feature.
\begin{figure} [ht]
\centerline{\includegraphics[height=7.5cm,angle=-90]{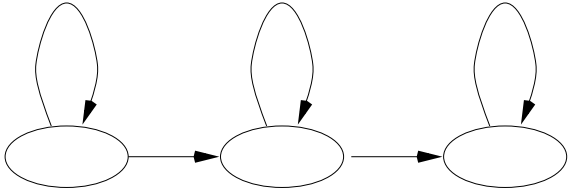}}
\caption{Topology of states used for each model of feature}
\label{fig:topology}
\end{figure}

Designing the right model in pattern recognition is known as
the model selection problem and is still an open area of
research. Based on our experience in speech recognition, we used
the well known left-right model (figure~\ref{fig:topology}),
which efficiently performs temporal segmentation of the
data. Recognition begins in the leftmost state, and each time an event
characterizing the feature is recognized it advances to the next state
to the right. When the rightmost state has been reached, the
recognition of the feature is complete.

The number of states is generally chosen as a monotone function of the
length of the pattern to be identified according to the state
duration probabilities. 

In the model depicted in 
figure   \ref{fig:topology}, the duration in state j may be defined as  :
\begin{eqnarray*}
d_j(0) & = & 0 \\
d_j(1) & = & a_{ijk}, ~~~~ i \neq j \neq k \\
d_j(n) & = & (1 - a_{ijk})\cdot a_{jjj}^{n-2}\cdot (1 - a_{jjj}), \\
& ~ & n ~\geq ~2.
\end{eqnarray*}
The state duration in a HMM2 is governed by two parameters:
the probability of entering a state only once, and the
probability of visiting a state at least twice, with the latter
modeled as a geometric decay.  This distribution fits a
probability density of durations \cite {crystal88a} better than the classical
exponential distribution of an HMM1.  This property is of great
interest in speech recognition when a HMM2 models a phoneme in which a
state captures only 1 or 2 frames.

This choice gives generally high rate of recognition. Sometimes,
adding or suppressing one or two states has been experimentally
observed to increase the rate of recognition. The number of states is
generally chosen to be the same for all the models.


\item Collecting and labeling a corpus of sequence of observations
during several runs to perform learning.


The corpus is used to adjust the parameters of the model to take into
account the statistical properties of the sequences of sensor
data. Typically, the corpus consists of a set of sequences of
features collected during several runs of the mobile robot. So, these
runs should be as representative as possible of the set of situations
in which features could be detected. The construction of the corpus is
time-consuming, but is crucial to effective learning. 
A model is trained with sequences of sensor data corresponding to the
particular feature it represents. Since a run is composed of a
sequence of features (and not only one feature), we need to segment
and label each run. To perform this operation, we use the previously
defined coarse rules to identify each feature and extract the relevant
sequences of data.  Finally, we
group the segments of the runs corresponding to the same feature to form a
corpus to train the model of that feature;

\item Defining a way to be able to detect all the features seen during
a run of the robot.

For this, the robot's environment is described by means of a grammar
that restricts the set of possible sequences of models. Using this
grammar, all the HMM2s are merged in a bigger HMM on which the Viterbi
algorithm is used. 
This grammar is a regular expression describing the legal sequences of HMM2s; it is used to know the possible ways of merging the HMM2s and their likelihood. More formally, this grammar represents all possible Markov chains corresponding to the hidden part of the merged models. In these chains, nodes correspond to HMM2s associated with a particular feature. Edges between two HMM2s correspond to a merge between the last state of one HMM2 and the first state of the other HMM2. The probability associated with each edge represents the likehood of the merge.

Then, the most likely sequence of states, as
determined by the Viterbi algorithm, determines the ordered list of
features that the robot saw during its run. It must be noted that the
list of models is known only when the run is completed. We make the
hypothesis that two or more of the features cannot overlap.
The use of a grammar has another important advantage.  It allows the
elimination of some sequences that will never happen in the
environment.  From a computational point of view, the grammar will avoid
some useless calculations.

The grammar can be given apriori or learned. To learn the grammar, we
use the former models and estimate them on unsegmented data like in
the recognition phase. Specifically, we merge all the models seen by the
robot during a complete run into a larger model corresponding to the
sequence of observed items and train the resulting model with the
unsegmented data. 

\item Evaluating the rate of recognition.

For this, we define a test corpus composed of several runs. For each
run, a human compares the sequence of features composing the run, using
knowledge of the environment, with what
has been detected by the Viterbi algorithm. A feature is recognized if
it is detected by the corresponding model close to its real geometric
position. A few types of errors can occur:
\begin{description}
\item[Insertion:] the robot has seen a non-existing feature (false positive). This
corresponds to an over-segmentation in the recognition process.
Insertions are currently considered when the width of the inserted
feature is more than 80 centimeters;
\item[Deletion:] the robot has missed the feature (false negative);
\item[Substitution:] the robot has confused the feature
with another.
\end{description}

In the experiments that we have run, the results are summarized first
as confusion matrices, where an element $c_{ij}$ is the number of
times the model $j$ has been recognized when the right answer was
feature $i$, and second with the global rate of recognition,
insertion, substitution and deletion.

\end{enumerate}
In the two following sections, we present two experiments where we
used second-order Hidden Markov Models to detect features using
sequence of mobile-robot sensor data. In each section, after a
brief description of the problem and the mobile robot used, we explain
the specific solution to each of the issues introduced in this section.

\section{First experiment: Learning and recognition of features in an indoor 
structured environment}
\label{sec:first-experiment}
In this first experiment, we used second order Hidden Markov Models to
learn and to recognize indoor features such as T-intersections and open
doors given sequences of data from  ultrasonic sensors of
an autonomous mobile robot.  These features are generally called
{\em places}. 


\subsection{The Nomad200 mobile robot}

\begin{figure} [ht]
\centerline{\includegraphics[width=5cm,angle=-90]{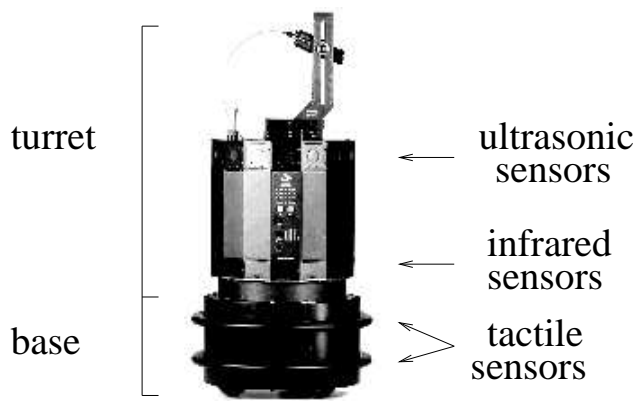}}
\caption{Our mobile robot}
\label{fig:gaston}
\end{figure}

In this experiment, we used a Nomad200 (figure~\ref{fig:gaston})
manufactured by Nomadic
Technologies\footnote{http://www.robots.com}. It is composed of a base
and a turret.  The base consists of 3 wheels and tactile sensors. The
turret is an uniform 16-sided polygon. On each side, there is an
infrared and an ultrasonic sensor.  The turret can rotate
independently of the base.

\paragraph{Tactile Sensors:}
A ring of 20 tactile sensors surrounds the 
base. They detect contact with objects. They 
are just used for emergency situations. They 
are associated with low-level reflexes such as 
emergency stop and backward movement.

\paragraph{Ultrasonic Sensors:}
The angle between two ultrasonic sensors is 22.5 degrees, and each
ultrasonic sensor has a beam width of approximately 23.6 degrees. By
examining all 16 sensors, we can obtain a 360 degree panoramic view
fairly rapidly. The ultrasonic sensors give range information from 17
to 255 inches.  But the quality of the range information greatly
depends on the surface of reflection and the angle of incidence
between the ultrasonic sensor and the object.

\paragraph{Infrared Sensors:}
The infrared sensors measure the light differences between an emitted
light and an reflected light. They are very sensitive to the ambient 
light, the object color, and the object orientation.
We assume that for short distances the 
range information is acceptable, so we just use 
infrared sensors for the areas shorter than 17 inches, where the
ultrasonic sensors are not usable.

\subsection{Specifics of HMM2 application to indoor place identification}
Here we discuss the specific issues arising from applying HMM2s to the
problem of indoor place identification, along with our solutions to
those issues.  The numbering corresponds to the numbering of the steps
in section \ref{sec:application-robotics}.

\subsubsection{The set of places}
\begin{figure}[htb]
\centerline{\includegraphics[width=7cm]{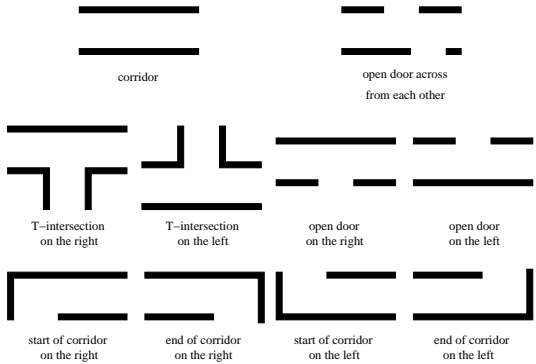}}
\caption{The 10 models to recognize}
\label{fig:10models}
\end{figure}
Currently, we model ten distinctive places that are representative of
an office environment: a corridor, a T-intersection on the right
(resp. left) of the corridor, an open door on the right (resp. left)
of the corridor, a ``starting'' corner on the right (resp. left) when
the robot moves away from the corner, an ``ending'' corner on the
right (resp. left) side of the corridor when the robot arrives at this
corner, two open doors across from each other
(figure~\ref{fig:10models}). This set of items is a complete
description of what the mobile 
robot can see during its run. All other unforeseen objects, like people
wandering along in a corridor, are treated as noise.

\begin{figure} [htbp]
\centerline{\includegraphics[height=5cm]{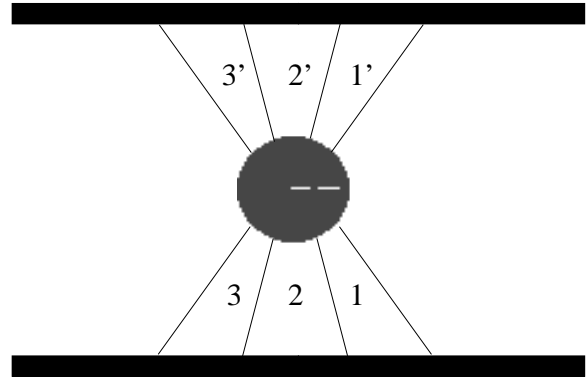}}
\caption{The six sonars used for the characterization on each side}
\label{fig:segmentation}
\end{figure}
\begin{figure} [htbp]
\centerline{\includegraphics[width=7cm,height=4cm]{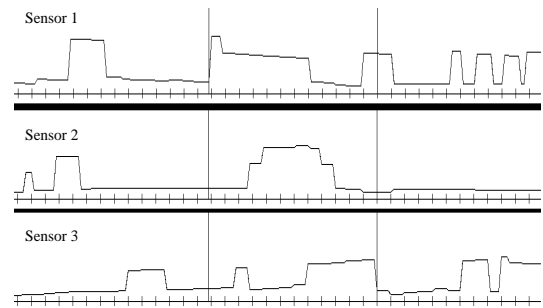}}
\caption{The characterization corresponding to a T-intersection on the
  right side of the robot} 
\label{fig:sensor}
\end{figure}
\begin{figure} [htbp]
\centerline{\includegraphics[width=7cm]{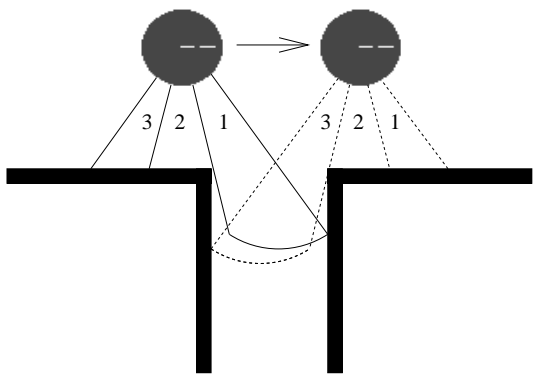}}
\caption{The three sonars used for the segmentation of a T-intersection}
\label{fig:T-intersection}
\end{figure}

To characterize each feature, we need to select the pertinent sensor
measures to observe a place. This task is complex because the sensor
measures are noisy and because at the same time that there is a place
on the right side of the robot, there is another place on the left
side of the robot. For these reasons, we choose to characterize
features separately for each side, using the sensors perpendicular to
each wall of the corridor and its two neighbor sensors
(figure~\ref{fig:segmentation}). These three sensors normally give
valid measures.
Since all places except the corridor 
cause a noticeable variation on these three sensors over time, we define the 
beginning of a place on one side when the first sensor's measure suddenly increases and the end of a place when the last sensor's measure suddenly
decreases. Figure~\ref{fig:sensor} shows an example of the
segmentation on the right side with these three sensors of a part of
an acquisition corresponding to a T-intersection. 
The first line segment is the
beginning of the T-intersection (sudden increase on the first sensor),
and the second line segment is the end of the T-intersection (sudden
decrease on the third sensor). To the left of the first line and
to the right of the second line are corridors.
Figure~\ref{fig:T-intersection} shows the position of the robot at the
beginning and at the end of the T-intersection and the measures of the
three sensors used at these two positions for the characterization. 
Next, we must 
define ``global places'' taking into account what can be seen on the
right side and on the left side simultaneously. To build the global
places, we combine the 5 previous places observable on the right side
with the 5 places observable on the left side.  

\begin{figure*} [htbp]
\centerline{\includegraphics[width=0.6\textwidth]{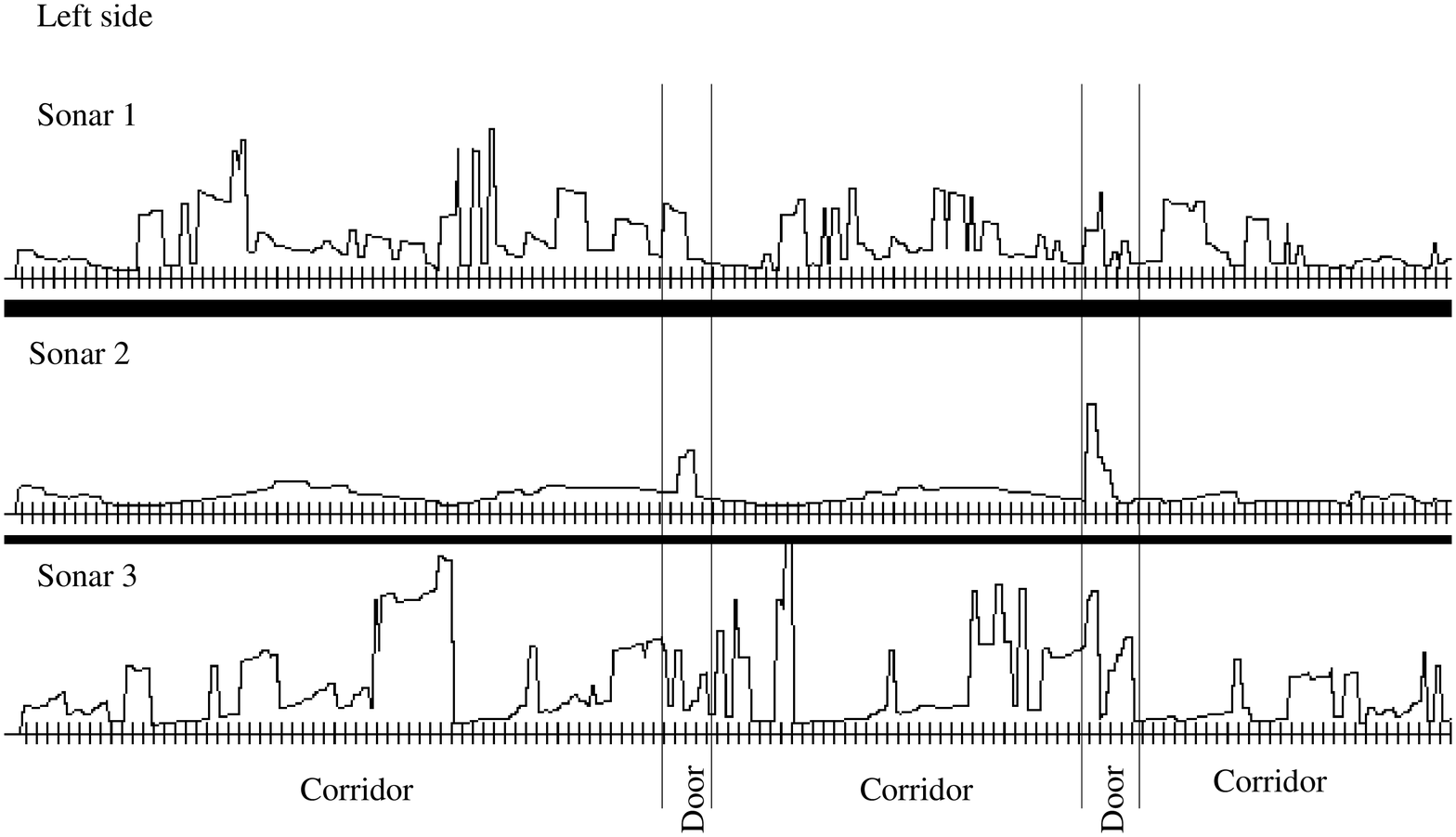}}
\centerline{\includegraphics[width=0.6\textwidth]{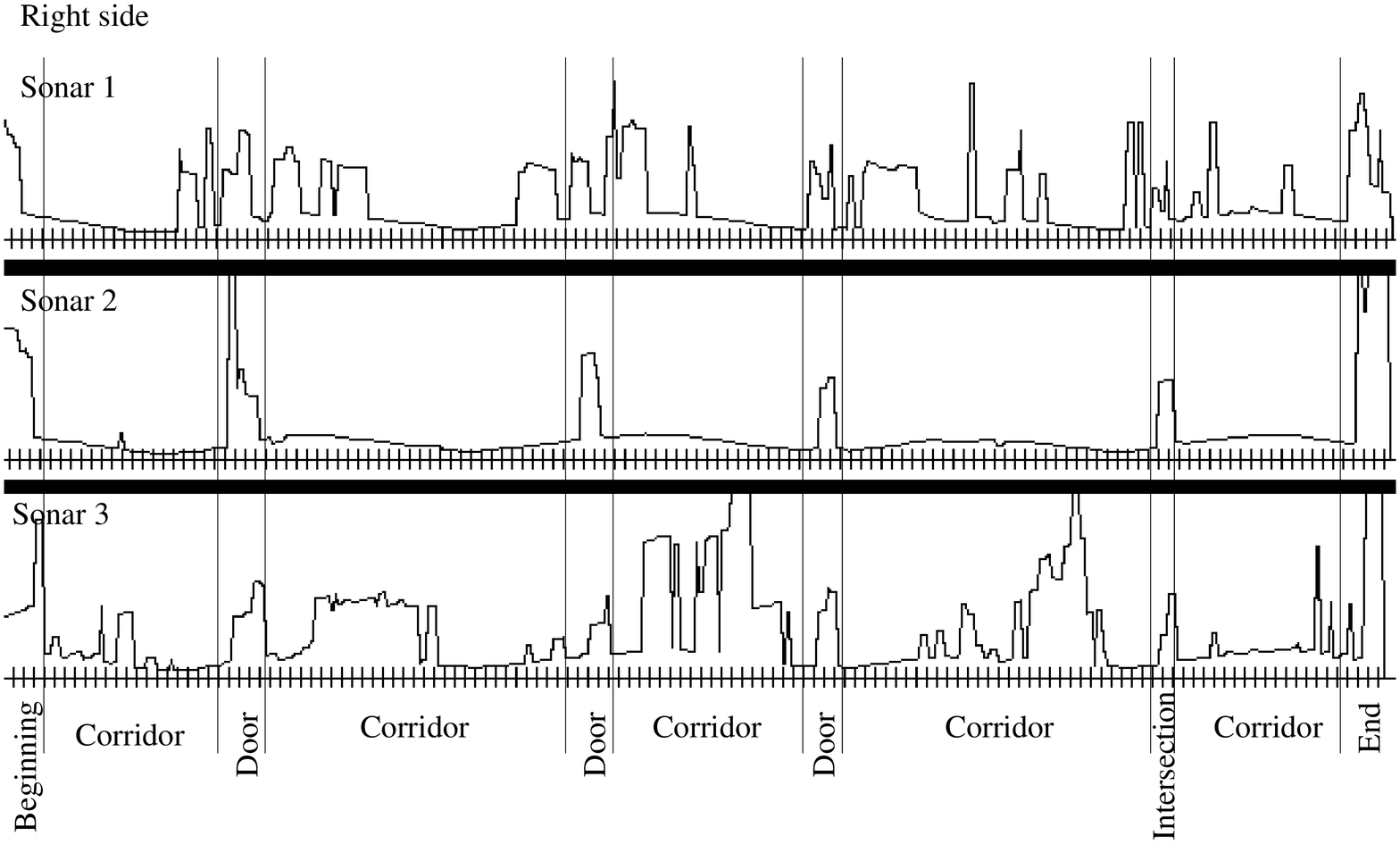}}
\centerline{\includegraphics[width=0.6\textwidth]{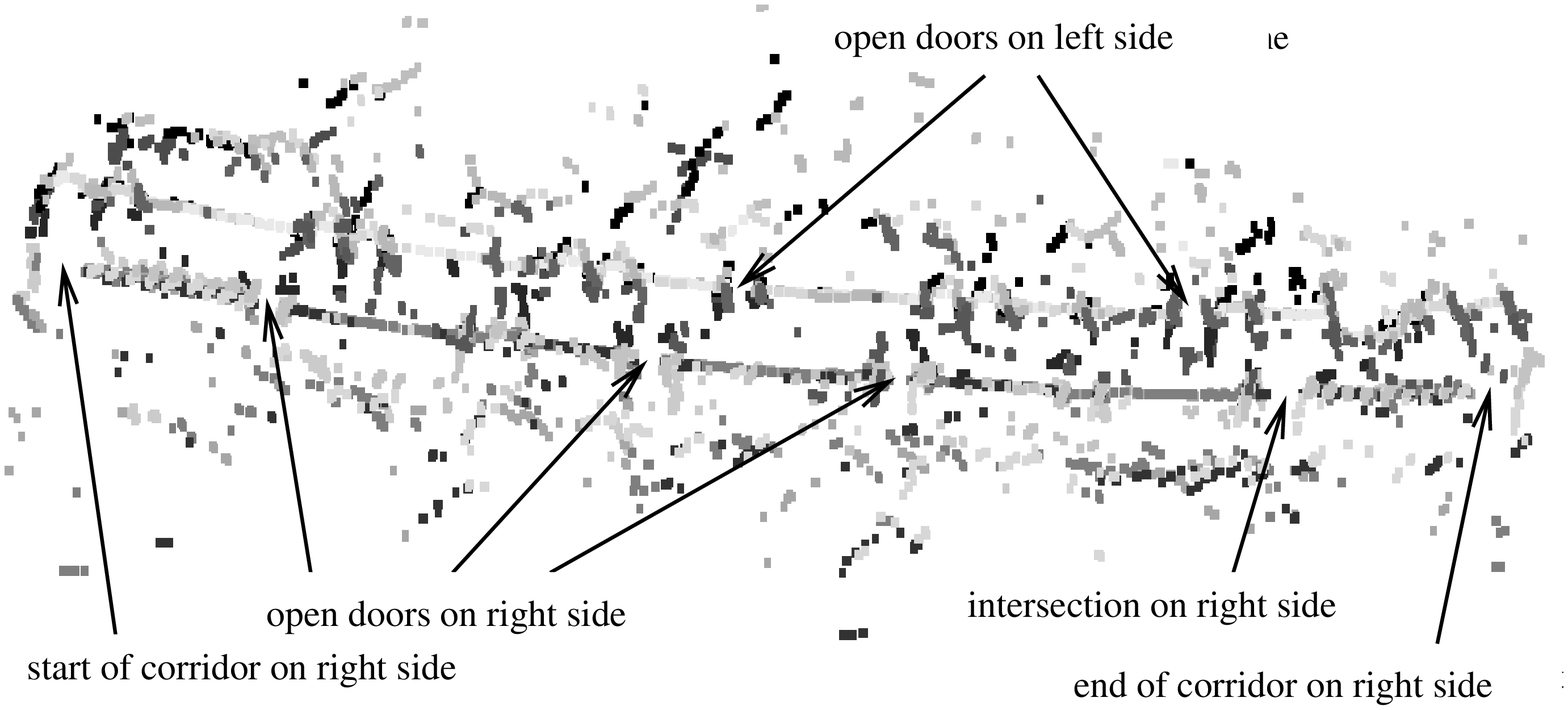}}
\caption{Example of characterization of the 10 places}
\label{fig:new-seg}
\end{figure*}

An example of the characterization of these 10 places is given in
figure~\ref{fig:new-seg}. This characterization will be used for
segmentation and labeling the corpus for training and evaluation. 

\subsubsection{The model to represent each place}
In the formalism described in section~\ref{sec:hmm2}, each place to be recognized is modeled by an HMM2 whose topology is depicted in figure~\ref{fig:topology}.

As the robot is equipped with 16 ultrasonic sensors, the HMM2 models the 16-dimensional, real-valued signal coming from the battery of ultrasonic sensors.  


\subsubsection{Corpus collecting and labeling}
\begin{figure} [ht]
  \centerline{\includegraphics[height=7.5cm,angle=-90]{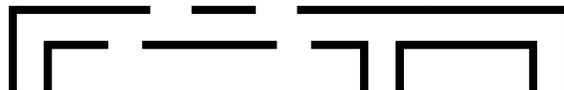}}
  \caption{The corridor used to make the learning corpus}
  \label{fig:corridor}
\end{figure}

We built a corpus to train a model for each of the 10 places. For this,
our mobile robot made 50 passes (back and forth) in a very long
corridor (approximately 30 meters).  This corridor
(figure~\ref{fig:corridor}) contains two corners (one at the start of
the corridor and one at the end), a T-intersection and some open doors
(at least four, and not always the same). The robot ran with a simple
navigation algorithm~\cite{Aycard97a} to stay in the middle of the
corridor in a direction parallel to the two walls constituting the
corridor. While running, the robot stored all of its ultrasonic sensor
measures. The acquisitions were done in real conditions with people
wandering in the lab, doors completely or partially opened and static
obstacles like shelves.

A pass in the corridor contains not only one place but all the places
seen while running in the corridor. To learn a particular place, we
must manually segment and label passes in distinctive places. The goal
of the segmentation and the labeling is to identify the sequence of
places the robot saw during a given pass. To perform this task, we use
the rules defined to characterize a place. Finally,
we group the segments from each pass corresponding to the same place.
Each learning
corpus associated with a model contains sequences of observations of the corresponding place.

\subsubsection{The recognition phase}
\label{sec:reco}
The goal of the recognition process is to identify the 9 places
 in the
corridor. We use a tenth model for the corridor because the Viterbi
algorithm needs to map each frame to a model during recognition.
The corridor model connects 2 items much like a silence between 2
words in speech recognition.  During this experiment, the robot uses
its own reactive algorithm to navigate in the corridor and must decide
which places have been encountered during the run.  We took 40
acquisitions and used the ten models trained to perform the
recognition. The recognition is independently processed on each side.

\subsection{Results and discussion}
Results are given in table~\ref{matrix1} and~\ref{matrix2}.
\begin{table*}[ht]
\centerline{
\begin{tabular}{||c||c|c|c|c|c|c|c|c|c|c||} \hline
             &  right &  right &  right &  right &  left  &  left  
             &  left  &  left  & door & Ins. \\
             & start  &  end   & inter. &  door  & start  &  end   
             & inter. &  door  &  door  & \\ \hline \hline
right start  &  7 &  0 &  0 &  0 &  0 &  0 &  0 &  0 &  0 &  0 \\ \hline
right end    &  0 &  7 &  0 &  0 &  0 &  1 &  0 &  1 &  0 &  0 \\ \hline
right inter. &  0 &  0 &  7 &  0 &  0 &  0 &  0 &  0 &  0 &  0 \\ \hline
right door   &  0 &  0 &  0 & 42 &  0 &  0 &  0 &  1 &  1 & 25 \\ \hline
left start   &  0 &  0 &  0 &  0 &  8 &  0 &  0 &  0 &  0 &  0 \\ \hline
left end     &  0 &  0 &  0 &  0 &  0 &  6 &  0 &  0 &  0 &  0 \\ \hline
left inter.  &  0 &  0 &  0 &  0 &  0 &  0 &  8 &  0 &  0 &  0 \\ \hline
left door    &  1 &  0 &  0 &  4 &  0 &  0 &  1 & 43 &  1 & 34 \\ \hline
door door    &  0 &  0 &  1 &  0 &  0 &  0 &  0 &  1 &  2 &  1 \\ \hline
deletions    &  0 &  1 &  0 &  0 &  0 &  0 &  0 &  0 &  0 &  0 \\\hline \hline
Total        &  8 &  8 &  8 & 46 &  8 &  7 &  9 & 46 &  4 & 60 \\ \hline
\% reco.     & 88 & 88 & 88 & 91 & 100& 86 & 89 & 93 & 50 &\\ \hline
\end{tabular}}
\caption{Confusion matrix of places}
\label{matrix1}
\end{table*}

\begin{table}[ht]
\centerline{
\begin{tabular}{||c||c||c||}        \hline
             & number & \%  \\ \hline
Seen       &    144 & 100 \\ \hline
Recognized &    130 &  90 \\ \hline
Substituted &     11 &   9 \\ \hline
Deleted    &      2 &   1 \\ \hline
Inserted   &     60 &  42 \\ \hline
\end{tabular}}
\caption{Global rate of recognition}
\label{matrix2}
\end{table}
We notice that the rate of recognition are very high, and the rate of confusion are very low. This is due to the fact that each place has a very particular pattern, and so it is very difficult to confuse it with an other. 
In fact, HMM2 used hidden characteristics (i.e, characteristics not explicitly given during the segmentation and the labelization of places) to perform discrimination between places. In particular, a place is characterized by variations on sensors on one side of the robot, but too with variations on sensors located on the rear or the front of the robot. Observations of sensors situated on
the front of the robot are very different when the robot is in the
middle of the corridor than at the end of the corridor. So, the models
of start of corridor (resp. end of corridor) could be recognized only
when observations of front and rear sensors correspond to the start of
a corridor (resp. the end of a corridor), which will rarely occur when
the robot is in the middle of the corridor. So, it is nearly
impossible to have insertions of the start of a corridor (resp. end of
corridor) in the middle of the corridor. 

HMM2 have been able to learn this type of hidden characteristics and to use them to perform discrimination during recognition.

But, we see that T-intersection and open doors have very similar characteristics using sensor information, and there is nearly no confusion between these two places. An other characteristic has been learned by the HMM2 to perform the discrimination between these two places. The width of open doors is different from the width of intersections, the discrimination between these two types of
places is improved because of the duration modeling capabilities of the HMM2,
as presented above and as shown by~\cite{Mari97}.

The rate of recognition of two open doors across from each other is
mediocre (50\%). There exists a great variety of
doors that can overlap and we only define one model that represents all these situations. So this model is a very general model of two doors across from each other. Defining more specific models of this place would lead to increase the associate rate of recognition.

The major problem is the high rate of insertion. Most of the insertions are due to the inaccuracy of the navigation algorithm
and to the unexpected obstacles. Sometimes the mobile 
robot has to avoid people or obstacles, and in these cases it does not
always run parallel to the two
walls, and in the middle of the corridor. These conditions cause
reflections on some sensors which are interpreted as places. A level incorporating 
knowledge about the environment should fix this problem. 

Finally, the global rate of recognition is 92\%. Insertions of places are 42\%.  Deletions are at a very low probability level  (less than 1.5\%).

\section{Second experiment: Situation identification for planetary rovers: Learning and Recognition}
\label{sec:second-experiment}
In a second experiment, we want to detect particular features (which
we call {\em situations}) when an outdoor teleoperated robot is exploring an unknown
environment.

This experiment has three main differences with the previous one:
\begin{enumerate}
\item the robot is an outdoor robot;
\item the sensors used as the observation are of a different type than in the indoor experiment;
\item we performed multiple learning and recognition scenarios using
different set of sensors. These experiments have been done to test the
robustness of the detection if some sensors break down.
\end{enumerate}

\subsection{Marsokhod rover}
\label{sec:marsokhod}
\begin{figure}[ht]
\centerline{\includegraphics[width=2.1in]{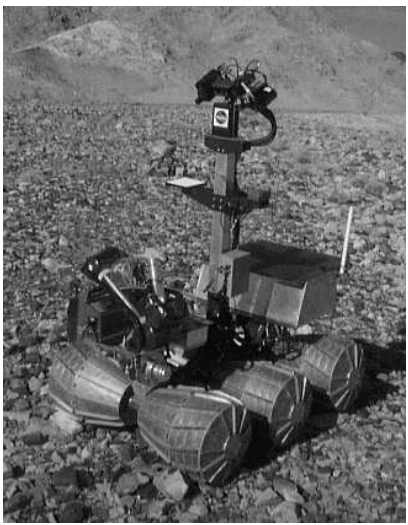}}
\caption{The Marsokhod rover}
\label{fig:marsokhod}
\end{figure}

The rover used in this experiment is a Marsokhod rover (see figure
\ref{fig:marsokhod}), a medium-sized planetary rover originally
developed for the Russian Mars exploration program; in the NASA
Marsokhod, the instruments and electronics have been changed from the
original.  The rover has six wheels, independently driven,\footnote{For
the experiments, the right rear wheel had a broken gear, so it rolled
passively.} with three chassis segments that articulate
independently.  It is configured with imaging cameras, a spectrometer,
and an arm.  The Marsokhod platform has been demonstrated at field
tests from 1993--99 in Russia, Hawaii, and deserts of Arizona and
California; the field tests were designed to study user interface
issues, science instrument selection, and autonomy technologies.

The Marsokhod is controlled either through sequences or direct tele-operation.
In either case the rover is sent discrete commands that describe motion in
terms of translation and rotation rate and total time/distance.
The Marsokhod is instrumented with sensors that measure body, arm, and
pan/tilt geometry, wheel odometry and currents, and battery currents.
The sensors that are used in this paper are roll (angle from vertical
in direction perpendicular to travel), pitch (angle from vertical in
direction of travel), and motor currents in each of the 6 wheels.

The experiments in this paper were performed in an outdoor
``sandbox,'' which is a gravel and sand area about 20m x 20m, with
assorted rocks and some topography.  This space is used to perform
small-scale tests in a reasonable approximation of a planetary (Martian)
environment.  We distinguish between the small (less than approx. 15cm
high) and large rocks (greater than approx. 15cm high).  We also
distinguish between the one large hill (approx. 1m high) and the three
small hills (0.3-0.5m high).

\subsection{Specifics of HMM2 application to outdoor situation identification}

Here we discuss the specific issues arising from applying HMM2s to the
problem of outdoor situation identification, along with our solutions to
those issues.  The numbering corresponds to the numbering of the steps
in section \ref{sec:application-robotics}.

\subsubsection{The set of situations}
Currently, we model six distinct situations that are representative of
a typical outdoor exploration environment: when the robot is climbing
a small rock on its left (resp. right) side, a big rock on its left
side,\footnote{The situation of a big rock on the right side was not
considered because of the non-functional right-side wheel.} a small
(resp. big) hill, and a default situation of level ground.

This set of items is considered to be a complete description of
what the mobile robot can see during its runs. All other unforeseen
situations, like flat rocks or holes, are treated as noise.

One possible application of this technique would be to identify
internal faults of the rover (e.g., broken encoders, stuck wheels).
This would require instrumenting the rover to cause faults on command,
which is not currently possible on the Marsokhod.  Instead, the
situations used in this experiment were chosen to illustrate the
possibility of using a limited sensor suite to identify situations,
and in fact some sensors were not used (such as joint angles) so that
the problem would become more challenging.

As Hidden Markov Models have the ability to model signals whose
properties change with time, we have to choose a set of sensors (as
the observation) that have noticeable variations when the Marsokhod is
crossing a rock or a hill.  From the sensors described in
section~\ref{sec:marsokhod}, we identified eight such sensors: roll,
pitch, and the six wheel currents.  We define coarse rules to identify
each situation (used by humans for segmentation and labeling the corpus
for training and evaluation):
\begin{itemize}
\item
When the robot crosses a small (resp. big) rock on its left, we
notice a distinct sensor pattern. In all cases, the roll
sensor shows a small (resp. big) increase when climbing the rock,
then a small (resp. big), sudden decrease when descending from
the rock.  These two variations usually appear sequentially on the
front, middle, and rear left wheels.  The pitch sensor always shows a
small (resp. big) increase, then a small (resp. big), sudden
decrease, and finally a small (resp. big) increase.  There is
little variation on the right wheels.
\item
When the robot crosses a small rock on its
right side, we observe variations symmetric to the case of a small rock
on the left side.
\item
When the robot crosses a small (resp. big) hill,
the pitch sensor usually shows a small (resp. big) increase, then

a small (resp. big) decrease, and finally a small (resp. big)
increase.  There is not always variation in the roll
sensor.  However, there is a gradual, small
(resp. big) increase followed by a gradual, small (resp. big)
decrease on all (or almost all) the six wheel current
sensors.
\end{itemize}

\subsubsection{The model to represent each situation}
\begin{figure} [ht]
\centerline{\includegraphics[width=\columnwidth]{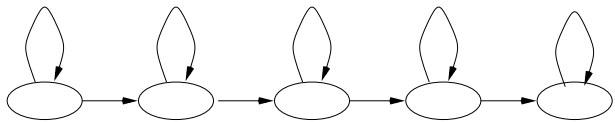}}
\caption{Topology of states used for each model of situation}
\label{fig:topology5}
\end{figure}

In the formalism described in section~\ref{sec:hmm2}, each situation
to be recognized is modeled by a HMM2 whose topology is depicted in
figure~\ref{fig:topology5}. This topology is well suited for the type
of recognition we want to perform. In this experiment, each model has
five states to model the successive events characterizing a particular
situation. This choice has been experimentally shown to give
the best rate of recognition.


\subsubsection{Corpus collecting and labeling}
\begin{figure*} [htbp]
\centerline{\includegraphics[width=\textwidth]{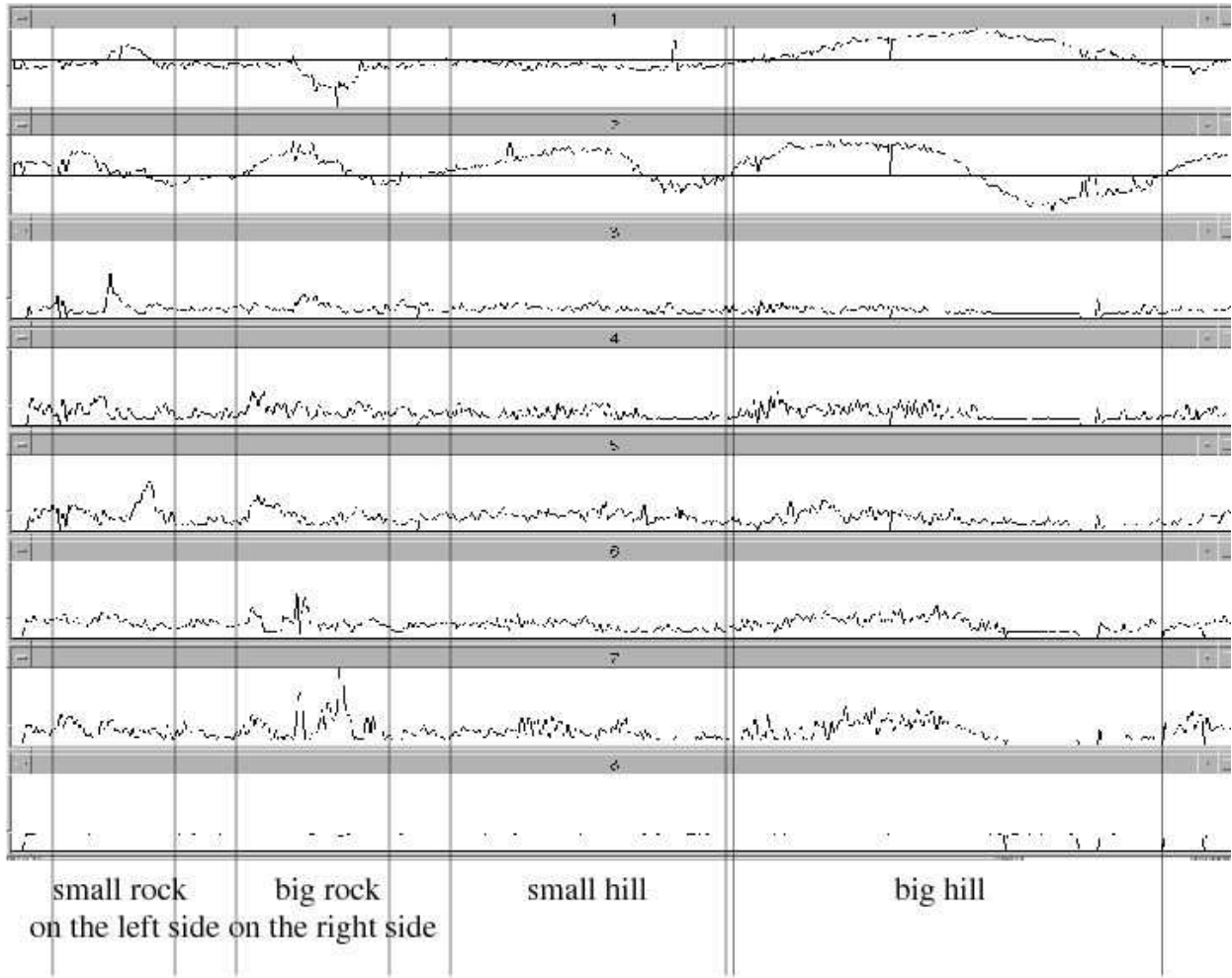}}
\caption{Segmentation and labeling of a run.}
\label{fig:segment-label}
\end{figure*}

We built six corpora to train a model for each situation. For this,
our mobile robot made approximately fifty runs in the sandbox. For
each run, the robot received one discrete translation command ranging
from three meters to twenty meters. Rotation motions are not part of
the corpus. 
Each run
contains different situations, but each run is unique (i.e., the area
traversed and the sequence of situations during the run is different
each time).  A run contains not only one situation but all the
situations seen while running.  For each run, we noted the situations
seen during the run, for later segmentation and labeling purposes.

The rules defined to characterize a situation are used to segment and label each run. An example of
segmentation and labeling is given in
figure~\ref{fig:segment-label}.  The sensors are in the following
order (from the top): roll, pitch, the three left wheel currents,
and the three right wheel currents.  A vertical line marks
the beginning or the end of a situation. The default
situation alternates with the other situations. The sequence of
situations in the figure is the following (as
labeled in the figure): small rock on the left side,
default situation, big rock on the right side, default situation,
small hill, default situation, and big hill.

\subsubsection{Model training}
In this experiment, we do not need to interpolate the observations
done by the robot, because it always moves at approximately the same
translation speed. As we want to compare different possibilities and
test if the detection is usable even if some sensors break down, we
train a separate model for each of three sets of input
data. The observations used as input of each model to train consist
of:
\begin{itemize}
\item eight coefficients: the first
derivative (i.e., the variation) of the values of the eight sensors
used for segmentation.
\item six coefficients: the first
derivative (i.e., the variation) of the values of the six wheel
current sensors.
\item two coefficients:  the first
derivative (i.e., the variation) of the values of the roll and the
pitch sensors.
\end{itemize}
Each training uses segmented data, and each model is trained
independently with its corpus.

There are two reasons for training three different models.
First is to check whether the eight sensors used for the segmentation are
necessary to learn and recognize situations, or whether a subset
is sufficient.  Second, we want to be able to recognize
situations even if one or more sensors do not work; e.g., if 
some wheel sensors do not work it will affect (during 
recognition) the models using the six wheel current sensors or the
eight sensors but not the models using just the roll and pitch
sensors.

\subsubsection{The recognition phase}

The goal of recognition is to identify the five situations
(small rock on the left or right; big rock on the left; 
small or big hill) while the robot moves in the
sandbox. The default situation model connects two items much like
silence between two words in speech recognition.

During the recognition phase, the robot was operated as for corpus
collecting.  We took approximately 40 acquisitions and used the six
trained models to perform the recognition.
We perform three independent recognitions, corresponding to the three
learning situations.

\subsubsection{Results and discussion}
In each confusion matrix, the acronyms used are: BL~=~big rock on the
left, SL~=~small rock on the left, SR~=~small rock on the right,
BH~=~big hill, and SH~=~small hill. The results of the three
independent experiments are shown and analyzed in the three next
subsections. In the fourth subsection, we present a global analysis of the
results.

\paragraph{Experiment with eight sensors}
\begin{table}[ht]
\centerline{
\begin{tabular}{|c||c|c|c|c|c||c|} \hline
        & BL    & SL    & SR    & BH    & SH    & Ins   \\ \hline
BL      & 19    & 3     & 1     & -     & -     & 9     \\ \hline 
SL      & 3     & 25    & -     & -     & -     & 12    \\ \hline 
SR      & 1     & 2     & 31    & -     & 1     & 26    \\ \hline 
BH      & 1     & -     & -     & 20    & 2     & 15    \\ \hline 
SH      & -     & -     & -     & 1     & 23    & 28    \\ \hline 
Del     & 1     & 1     & -     & -     & -     & -     \\ \hline \hline
Total   & 25    & 31    & 32    & 21    & 26    & 90    \\ \hline
Reco    & 76\%  & 81\%  & 97\%  & 95\%  & 88\%  &       \\ \hline 
\end{tabular}
}
\caption{Confusion matrix of situations, eight sensors.}
\label{tab:confusion8}
\end{table}

\begin{table}[ht]
\centerline{
\begin{tabular}{|c||c|c|} \hline
               & number        & \%    \\ \hline
Seen           & 135           & 100   \\ \hline
Recognized     & 118           & 87    \\ \hline
Substituted    & 15            & 11    \\ \hline
Deleted                & 2             & 2     \\ \hline
Inserted       & 90            & 67    \\ \hline
\end{tabular}
}
\caption{Global rate of recognition, eight sensors.}
\label{tab:global8}
\end{table}


For eight sensors, as each situation can be easily distinguished from
the others, the global rate of recognition is excellent (87\%) (see
tables~\ref{tab:confusion8},~\ref{tab:global8}).  Small
(resp. big) rocks on the left are sometimes confused with big
(resp. small) rocks on the left; the signal provided by the
sensors does not contain the information necessary to discriminate these two
models.  In fact, the variations on the sensors are nearly the same.
The only criterion which distinguishes these two models is the
amplitude of the variation on the three left wheels, and visibly it is
not sufficient.  The small rocks on the right are perfectly
recognized.  This situation has a very distinctive pattern, and only
with difficulty can it be confused with another.  The fact that we
could not learn and recognize a situation where the robot is crossing
a big rock on its right avoids any confusion. 

The major problem is the high rate of insertion.  This rate is due to
the noise of the sensors being recognized as a situation. This is especially
the case for situations characterized only by small variations on a
part (or all) of the set of sensors, in particular the crossing of a
small hill.

\paragraph{Experiment with six sensors}
\begin{table}[ht]
\centerline{
\begin{tabular}{|c||c|c|c|c|c||c|} \hline
        & BL    & SL    & SR    & BH    & SH    & Ins \\ \hline
BL      & 17    & 5     & 1     & -     & -     & 10 \\ \hline
SL      & 4     & 24    & 2     & -     & 1     & 19 \\ \hline
SR      & 3     & -     & 29    & -     & 1     & 44 \\ \hline
BH      & -     & 1     & -     & 20    & 1     & 19 \\ \hline
SH      & 1     & -     & -     & 1     & 23    & 32 \\ \hline
Del     & -     & 1     & -     & -     & -     & - \\ \hline \hline
Total   & 25    & 31    & 32    & 21    & 26    & 124 \\ \hline
Reco    & 68\%  & 77\%  & 91\%  & 95\%  & 88\%  &       \\ \hline
\end{tabular}
}
\caption{Confusion matrix of situations, six sensors.}
\label{tab:confusion6}
\end{table}

\begin{table}[ht]
\centerline{
\begin{tabular}{|c||c|c|} \hline
               & number        & \%    \\ \hline
Seen           & 135           & 100   \\ \hline
Recognized     & 113           & 84    \\ \hline
Substituted    & 21            & 15    \\ \hline
Deleted                & 1             & 1     \\ \hline
Inserted       & 124           & 92    \\ \hline
\end{tabular}
}
\caption{Global rate of recognition, six sensors.}
\label{tab:global6}
\end{table}

With six sensors, the global rate of recognition is still very good
(see tables~\ref{tab:confusion6},~\ref{tab:global6}).
There is only four more percent of substitutions due to the loss of
information used to distinguish situations.  On the other hand, the
rate of insertion increased by 25\%.  With only the six wheel current
sensors, nearly one recognition out of two is an insertion. The six
wheel current sensors are very noisy, and the roll and pitch sensors
are useful to distinguish between simple noise and real situations.
This explains the increase of the insertions.

\paragraph{Experiment with two sensors}
\begin{table}[ht]
\centerline{
\begin{tabular}{|c||c|c|c|c|c||c|} \hline
        & BL    & SL    & SR    & BH    & SH    & Ins \\ \hline
BL      & 15    & 4     & -     & 6     & 1     & 1\\ \hline
SL      & 2     & 17    & 1     & -     & 9     & 15\\ \hline
SR      & 2     & 1     & 27    & 1     & 5     & 8\\ \hline
BH      & 5     & -     & -     & 14    & 2     & 6\\ \hline
SH      & -     & 7     & 4     & -     & 9     & 12\\ \hline
Del     & 1     & 2     & -     & -     & -     & -\\ \hline \hline
Total   & 25    & 31    & 32    & 21    & 26    & 42\\ \hline
Reco    & 60\%  & 55\%  & 84\%  & 67\%  & 35\%  & \\ \hline
\end{tabular}
}
\caption{Confusion matrix of situations, two sensors.}
\label{tab:confusion2}
\end{table}

\begin{table}[ht]
\centerline{
\begin{tabular}{|c||c|c|} \hline
               & number        & \%    \\ \hline
Seen           & 135           & 100   \\ \hline
Recognized     & 82            & 61    \\ \hline
Substituted    & 50            & 37    \\ \hline
Deleted                & 3             & 2     \\ \hline
Inserted       & 42            & 31    \\ \hline
\end{tabular}
}
\caption{Global rate of recognition, two sensors.}
\label{tab:global2}
\end{table}

With only the roll and pitch sensors, the global rate of recognition
remains good, and the rate of insertions significantly decreases (see
tables~\ref{tab:confusion2},~\ref{tab:global2}).  In
fact, these two sensors are not too noisy, and when there is a
variation on these sensors it generally corresponds to a real
situation.  But these two sensors do not provide sufficient
information to distinguish between situations, which is why there is
a high rate of substitution.

\paragraph{Global analysis}
From the
results of experiments, we can draw some conclusions.  The best way to
perform recognition is with eight sensors: the rate of recognition
is a little bit better than for six sensors and the rate of insertion
is very smaller.  This can be explained by the fact that the six
wheels current sensors are very noisy, and the use of the roll and
pitch sensors, which are not too noisy, can distinguish
between a situation to recognize and a simple noise on the current
wheel sensors.  Nonetheless, the models learned in the two last
experiments could be useful in long exploration where sensors can fail,
since they provide usable, albeit less reliable, recognition.

This experiment can be  extended to fault detection, for example
broken wheels or sensor failure.  In fact, we can build one model of
a particular situation where all sensors work and several models
of this situation where one or several sensors are broken: for example a
model of a big rock on the right side and a model of a big rock on the
right when the front left wheel is broken.  Using these models, we can
recognize situations associated with the state of the sensors of the
robot, and detect failing of sensors or motors.




\section{Related work}
\label{sec:related-work}
A variety of approaches to state estimation and fault diagnosis have
been proposed in the control systems, artificial intelligence, and
robotics literature.

Techniques for state estimation of continuous values, such as Kalman
filters, can track multiple possible hypotheses \cite{rauch94,willsky76}.
However, they must be given an {\it a priori} model of each possible
state.  One of the strengths of the approach presented in this paper
is its ability to construct models from training data and then use them
for state identification.

Qualitative model-based diagnosis techniques
\cite{dekleer87,muscettola98} consider a snapshot of the system
rather than its history.  In addition, the system state is assumed
to be consistent with a propositional description of one of a set of
possible states. The presence of noisy data and temporal patterns
negates these assumptions.

Hidden Markov Models have been applied to fault detection in
continuous processes \cite{smyth94}; the model structure is
supplied, with only the transition probabilities learned from data.
In the approach in this paper, the HMM learns without prior knowledge
of the models.

Markov models have been widely used in mobile
robotics. Thrun~\cite{Thrun2001} reviews techniques based on
Markov models for three main problems in mobile robotics:
localization, map building and control. In these techniques, a Markov
model represents the environment, and a specific algorithm is used to
solve the problem. Our approach is different in a number of ways. We address
a different problem: the interpretation of temporal sequences of
mobile-robot sensor data to automatically detect
features. Moreover, we use very little {\it a priori} knowledge: in particular,
the topology of the model reflecting the human's understanding about
sequences of sensor data characterizing a particular feature. All the
other parameters of the model are estimated by learning. On the
contrary, the techniques presented in~\cite{Thrun2001} need some {\it
preliminary knowledge}: a map of the environment, a sensor model and
an actuator model. Usually, there is no learning component in these
techniques.

The most well-known work including a learning component is
by Koenig and Simmons~\cite{Koenig96}. They start with an {\it a priori} topological map
that is translated into a Markov model before any navigation takes
place. An extension of the Baum-Welch algorithm reestimates the Markov
model representing the environment, the sensor and actuator
models. There are a number of differences with this work:
\begin{itemize}
\item They use a Markov model to model the environment, whereas we use a
Markov model to model the sequence of events composing a particular
feature;
\item They need some {\it a priori} knowledge: a topological map of
the environment, and sensor and actuator models;
\item They make hypotheses on the value of some parameters to reduce
the number of parameters to estimate; we do not make any such hypothesis;
\item The observations they use are discrete, symbolic and
unidimensional. There are obtained by an abstraction (based on some
hypothesis) of the raw data of several sensors. Discrete symbolic and
unidimensional observations are the result of our method. They are
obtained by interpretation of a sequence of raw data from several sensor
without any prior hypothesis. 
\end{itemize}
Our work can be seen as a preliminary step for all of the work
presented in~\cite{Thrun2001}. 
We have previously built a sensor model based on the recognition rates
reported in this article; the model allowed robust localization in dynamic
environments~\cite{Aycard98}.


Hidden Markov Models have been used for 
interpretation of temporal sequences in robotics \cite{Hannaford91,Yang94}.
The approach 
presented in this paper is more robust for the following reasons:
\begin{itemize}
\item Yang, Xu, and Chen~\cite{Yang94} make some restrictions and
hypotheses on the observations they used: each component of the
observation is discretized, since he uses a HMM with discrete
observations. Moreover, each component of the observation
is presumed independent from the other. In our work, the probability of an
observation given a particular state is represented by a mixture of
Gaussians.  Thus we are able to deal with observations constituted by
noisy continuous data of different types\footnote{In the second
experiment, the observation is composed of three types of sensors.} of
sensors without any {\it a priori} assumption about the
independence of these data and without any discretization of the
data;
\item The particular approach we use is the second-order HMM (HMM2).  HMM2s
have been shown to be effective models to capture temporal variations
in speech~\cite{Mari97}, in many cases surpassing first-order
HMMs when the trajectory in the state space has to be accounted for.
For instance in the first experiment, due to the duration-modeling
capabilities of HMM2, the Viterbi algorithm was able to distinguish
an open door from a T-intersection. 
\end{itemize}

\section{Conclusion and future directions}
\label{sec:conclusion}
In this paper, we have presented a new method to learn to
automatically detect features for mobile robots using second-order
Hidden Markov Models. This method gives very good results, and has a
good robustness to noise, verifying that HMM2s are well suited for
this task. We showed that the process of recognition is robust to
dynamic environment. Features are detected even if they are quite
different from learned features: for instance, an open door is
recognized even if it is completely or partially opened. Moreover,
features are detected even if they are seen from a different point of
view. For instance, in contrast to Kortenkamp et
al~\cite{Kortenkamp92}, features are detected even if the robot is
not at a given distance from a wall and doesn't move in a direction
perfectly parallel to the two walls constituting the
corridor. Finally, our approach has been successfully tested in an
outdoor environment.

The results can be improved by adding more models to decrease the
intra-class variability (especially for open doors across from each
other) and to take into account contextual information. Another
criterion that could improve the results is to choose a
different number of states for each feature.

Moreover, the method takes advantage of analytical methods and
pattern classification methods. First, we analyze the
sensor data and define a model to represent the patterns in the data.
Secondly, the
learning algorithm automatically adjusts the parameters of the model
using a learning corpus.  Moreover, the learning
algorithm was able to extract more complex characteristics of a
feature than simple variations of sensor data between two consecutive
moments. For instance:
\begin{itemize}
\item The length of a sequence\footnote{the number of observations composing the sequence} of observations was taken into account in
the first experiment to detect the difference between a T-intersection
and an open door;
\item In the first experiment, the gradual decrease
(resp. increase) of the value of sensors located in front (resp. in
the rear) of the robot during time has been used to characterize a
start (resp. an end) of corridor;
\item The algorithm can find correlation between data from sensors of
different types to characterize a feature. For example, the
correlation of the roll, pitch and wheel current sensors is used to
characterize a situation in the second experiment.
\end{itemize}

However, our method has two drawbacks:
\begin{itemize}
\item As in Kortenkamp et al~\cite{Kortenkamp92}, 
a feature can only be recognized when it has been completely
visited. For example, the
robot would have to go back to turn at a T-intersection after it had
recognized it. 
\item Moreover, using the current technique, the list of places is
known only when the run has been completed.  To detect features online
during navigation, we can use a variant of the Viterbi algorithm
called Viterbi-block~\cite{Kriouile90a}. This algorithm is based on a
local optimum comparison of the different probabilities computed by
the Viterbi algorithm during time-warping of a shift-window of fixed
length in the signal and the different HMMs.  This algorithm can
detect features a few meters after they have been seen. We have used this
algorithm to perform localization in dynamic
environment~\cite{Aycard98}.
\end{itemize}

\bibliography{ijars}

\begin{thebibliography}{10}

\bibitem{Aycard97a}
O.~Aycard, F.~Charpillet, and J.-P. Haton.
\newblock A new approach to design fuzzy controllers for mobile robots
  navigation.
\newblock In {\em proceedings of IEEE International Symposium on Computational
  Intelligence in Robotics and Automation}, pages 68--73, 1997.

\bibitem{Aycard98}
O.~Aycard, P.~Laroche, and F.~Charpillet.
\newblock Mobile robot localization in dynamic environment using place
  recognition.
\newblock In {\em Proc. of ICRA'98}, 1998.

\bibitem{aycard04a}
Olivier Aycard, Jean-Fran\c{c}ois Mari, and Richard Washington.
\newblock Learning to automatically detect features for mobile robots using
  second-order hidden markov models.
\newblock {\em International Journal of Advanced Robotic Systems},
  4(1):231--245, Dec 2004.

\bibitem{Becker95}
C~Becker, J~Salas, K~Tokusei, and J-C Latombe.
\newblock Reliable navigation using landmarks.
\newblock In {\em IEEE Int. Conf. on Robotics and Automation}, 1995.

\bibitem{Borenstein96}
J.~Borenstein, B.~Everett, and L.~Feng.
\newblock {\em Navigating Mobile Robots: Systems and Techniques}.
\newblock A. K. Peters, Ltd., Wellesley, MA, 1996.

\bibitem{crystal88a}
T.H. Crystal and A.S. House.
\newblock {Segmental Durations in Connected Speech Signals: Current Results}.
\newblock {\em Journal of Acoustical Society of America}, 83(4):1553 -- 1573,
  April 1988.

\bibitem{dekleer87}
J.~de~Kleer and B.~C. Williams.
\newblock Diagnosing multiple faults.
\newblock {\em Artif. Intelligence}, 32:100--117, 1987.

\bibitem{Elfes89}
A~Elfes.
\newblock Using occupancy grids for mobile robot perception and navigation.
\newblock {\em IEEE Computer}, 22(6):46--57, June 1989.

\bibitem{Forney73}
G.D. Forney.
\newblock {The Viterbi Algorithm}.
\newblock {\em IEEE Transactions}, 61:268--278, March 1973.

\bibitem{Hannaford91}
B.~Hannaford and P.~Lee.
\newblock Hidden markov model analysis of force/troque information in
  telemanipulation.
\newblock {\em The International journal of robotics research}, October 1991.

\bibitem{Yang94}
C.~Chen J.~Yang, Y.~Xu.
\newblock Hidden markov model approach to skill learning and its application to
  telerobotics.
\newblock {\em IEEE transactions on robotics and automation}, October 1994.

\bibitem{Koenig96}
S.~Koenig and R.G. Simmons.
\newblock Unsupervised learning of probabilistic models for robot navigation.
\newblock In {\em proceedings of the International Conference on Robotics and
  Automation}, pages 2301--2308, 1996.

\bibitem{Kortenkamp92}
D~Kortenkamp, L~Douglas~Baker, and T~Weymouth.
\newblock Using gateways to build a route map.
\newblock In {\em proceedings of the 1992 IEEE International Conference on
  Intelligent Robots and Systems}, 1992.

\bibitem{Kriouile90a}
A.~Kriouile, J.-F. Mari, and J.-P. Haton.
\newblock {Some Improvements in Speech Recognition Algorithms based on HMM}.
\newblock In {\em Proceedings IEEE ICASSP'90 (International Conference
  Acoustics Speech and Signal Processing)}, pages 545 -- 548, Albuquerque,
  April 1990.

\bibitem{Kuipers00}
B.~Kuipers.
\newblock The spatial semantic hierarchy.
\newblock {\em Artificial Intelligence}, 2000.

\bibitem{Lazanas95}
A~Lazanas and J-C Latombe.
\newblock Landmark-based robot navigation.
\newblock {\em Algorithmica}, 1995.

\bibitem{Mari97}
J.-F. Mari, J.-P. Haton, and A.~Kriouile.
\newblock {Automatic Word Recognition Based on Second-Order Hidden Markov
  Models}.
\newblock {\em IEEE Transactions on Speech and Audio Processing}, 5, January
  1997.

\bibitem{muscettola98}
N.~Muscettola, P.~P. Nayak, B.~Pell, and B.~C. Williams.
\newblock Remote agent: To boldly go where no {AI} system has gone before.
\newblock {\em Artificial Intelligence}, 103(1/2), August 1998.

\bibitem{Rabiner89}
L.~R. Rabiner.
\newblock {A Tutorial on Hidden Markov Models and Selected Applications in
  Speech recognition}.
\newblock {\em IEEE Trans. on ASSP}, 77(2):257 -- 286, February 1989.

\bibitem{rauch94}
H.~E. Rauch.
\newblock Intelligent fault diagnosis and control reconfiguration.
\newblock {\em IEEE Ctrl. Sys.}, 14(3), 1994.

\bibitem{smyth94}
P.~Smyth.
\newblock Hidden {Markov} models for fault detection in dynamic systems.
\newblock {\em Pattern Recognition}, 27(1):149--164, January 1994.

\bibitem{Thrun98}
S.~Thrun.
\newblock Bayesian landmark learning for mobile robot localization.
\newblock {\em Machine Learning}, 1998.

\bibitem{Thrun2001}
S.~Thrun.
\newblock Probabilistic algorithms in robotics.
\newblock {\em AI Magazine}, 2001.

\bibitem{willsky76}
A.~S. Willsky.
\newblock A survey of design methods for failure detection in dynamic systems.
\newblock {\em Automatica}, 12:601--611, 1976.

\bibitem{Yamauchi95}
B~Yamauchi.
\newblock {\em Exploration and spatial learning in dynamic environnements}.
\newblock PhD thesis, Case Western Reserve University, 1995.

\end{thebibliography}

\end{document}